\documentclass{article}

\usepackage[preprint, nonatbib]{neurips_2019}

\usepackage[utf8]{inputenc} 
\usepackage[T1]{fontenc}    
\usepackage{hyperref}       
\usepackage{url}            
\usepackage{booktabs}       
\usepackage{amsfonts}       
\usepackage{amsmath}
\usepackage{nicefrac}       
\usepackage{microtype}      
\usepackage{graphicx}
\usepackage{booktabs}
\usepackage{multirow}
\usepackage{siunitx}

\usepackage{hyperref}

\usepackage{subfig}

\usepackage{algorithm}
\usepackage{algorithmic}

\renewcommand{\arraystretch}{1.2}

\title{An empirical study of pretrained representations for few-shot classification}

\author{%
  Tiago Ramalho, Thierry Sousbie, Stefano Peluchetti \\
  Cogent Labs\\
  Tokyo, Japan\\
  \texttt{\{tramalho, tsousbie, speluchetti\}@cogent.co.jp} \\
}

\expandafter\def\expandafter\normalsize\expandafter{%
    \normalsize
    \setlength\abovedisplayskip{4pt}
    \setlength\belowdisplayskip{4pt}
    \setlength\abovedisplayshortskip{4pt}
    \setlength\belowdisplayshortskip{4pt}
}

\begin{document}

\maketitle

\begin{abstract}
Recent algorithms with state-of-the-art few-shot classification results start their procedure by computing data features output by a large pretrained model. In this paper we systematically investigate which models provide the best representations for a few-shot image classification task when pretrained on the Imagenet dataset. We test their representations when used as the starting point for different few-shot classification algorithms. We observe that models trained on a supervised classification task have higher performance than models trained in an unsupervised manner even when transferred to out-of-distribution datasets. Models trained with adversarial robustness transfer better, while having slightly lower accuracy than supervised models. 
\end{abstract}

\section{Introduction}

Deep learning systems achieve remarkable performance for several classification problems when given large enough datasets \cite{krizhevsky2012imagenet,sun_revisiting_2017}. In the case of small amounts of training data, these models can easily overfit as they contain a very large number of parameters~\cite{bengio2012deep}. There are two main techniques to address these issues: if we have a medium-sized dataset, we can fine-tune the weights of a model trained on a larger dataset~\cite{kornblith_better_2018}; for very small datasets we rely on methods which learn from the 'training' dataset without multiple steps of gradient descent over all parameters in the model, known as meta-learning or few-shot classification~\cite{ravi_optimization_2016}.

Just as in the fine-tuning regime, in the few-shot classification case we also want to leverage large datasets for better final performance. This is almost always achieved by feeding the data representation produced by a deep neural network as input to the algorithm. For example, in the case of image classification practitioners will use a deep residual convolutional network pre-trained on a larger dataset to compute the features~\cite{rusu_meta-learning_2018,snell_prototypical_2017}.

Intuitively, if the representations produced by the pre-trained model are more discriminative of the different classes under consideration at test time, it will be easier for a few-shot classification method to produce better results~\cite{rusu_meta-learning_2018}. Previous work has shown that deeper convolutional models have higher accuracy and their feature quality appears to be the limiting factor in performance of most few-shot classification algorithms~\cite{chen_closer_2019}.

In this paper we will perform a systematic exploration of whether deep convolutional networks pretrained on the ImageNet dataset without few-shot classification in mind can transfer well to this task. Previous work has done a similar exploration for the case of fine-tuning~\cite{kornblith_better_2018}, but systematic reviews of few-shot classification methods have not looked at the transfer properties of pretrained models, instead focusing on the adaptation algorithm or architectural choices~\cite{chen_closer_2019,triantafillou_meta-dataset:_2019}.

After evaluating multiple models for 14 different datasets, we have come to the following conclusions:

\begin{itemize}
    \item Models pre-trained on a supervised classification task on the ImageNet dataset transfer well to other natural image datasets, but poorly to other types of image (e.g. MNIST, SVHN). The more data models are trained on, the better their transfer performance.
    \item Models trained with adversarial robustness~\cite{xie_feature_2018,ilyas_adversarial_2019} suffer a smaller performance decrease when transferred to out-of-distribution datasets. Their absolute performance is slightly lower than non-robust models.
    \item Models trained with an unsupervised learning loss~\cite{henaff_data-efficient_2019,bachman_learning_2019} do not reach the performance of models trained in a supervised manner, and do not seem to transfer better.
    \item Unlike what was reported in~\cite{snell_prototypical_2017} for simpler architectures trained from scratch, similarity-based few-shot classification methods used with pretrained models perform best when the similarity function is based on cosine similarity.
\end{itemize}

\section{Related work}

The earliest research in the field of few-shot classification focused on embedding space similarity based methods such as Matching~\cite{vinyals_matching_2016} and Prototype networks~\cite{snell_prototypical_2017}. These models are trained on  permutations of the meta-training dataset for a number of epochs, and then tested on unseen classes from the same distribution. Similarity based methods can be augmented with a decoder on top of the retrieved features to provide more flexibility in cases where the required output isn't a direct class mapping \cite{garnelo_conditional_2018,ramalho_adaptive_2019}.

MAML is a gradient-descent based method for metalearning~\cite{finn_model-agnostic_2017}, where the network is trained such that its weights converge to a configuration from which the model can be finetuned in very few steps of gradient descent on the support set. Recent work has demonstrated state of the art results for gradient-based methods \cite{rusu_meta-learning_2018}. In addition, MAML and related models have been shown to work particularly well for reinforcement learning scenarios~\cite{nagabandi_learning_2018}.

A third way of doing meta learning is to train a recurrent neural network to iterate over the whole support set and the new query point~\cite{ren_meta-learning_2018,mishra_simple_2017}. These methods require fewer assumptions on the structure of the activation space, but are more computationally intensive.

Most literature on metalearning tends to use the same distribution for the meta training set and the testing set, only holding out specific images or specific classes from training. This is a weaker form of generalization than cross-distribution generalization, which has recently been tested in \cite{triantafillou_meta-dataset:_2019}, and is the setting in which we are interested. The authors of this paper show that the performance of all metalearning methods degrades significantly when tested on a distribution significantly different to the one of the meta training set; and that a model which has been meta-trained on a number of different datasets generalizes better.

The authors of \cite{kornblith_better_2018} study transfer learning in the setting of models pretrained on imagenet. They show that models with better final accuracy on imagenet classification tend to have better transfer learning performance, which is encouraging for their application in few-shot classification.

The authors of Meta-Dataset explore the performance of few-shot classification methods when transferred across datasets~\cite{triantafillou_meta-dataset:_2019}. The authors survey a large number of state-of-the-art adaptation methods across a large variety of datasets with a convolutional backbone trained from scratch. They conclude that a model trained on a large variety of datasets has better transfer performance than a model trained exclusively on ImageNet. Their comparison across multiple adaptation methods leads us to conclude that quality of the features produced by a backbone affects performance much more strongly than the particular adaptation method. Unlike our work, they do not consider pretrained models and instead train a comparatively shallow backbone (ResNet-18) from scratch.

In~\cite{chen_closer_2019} we find another survey of few-shot classification methods. Here the authors measure the effect of convolutional backbone depth as well as classification algorithm on final performance. In this study only two datasets are considered, CUB and mini-ImageNet. As before, the authors find that deeper backbones provide better performance, but again restrict their analysis to shallow models trained from scratch.

Given the thorough analyses in the two papers mentioned above, we are encouraged to study the use of very deep convolutional networks trained on large datasets as feature extractors, as that appears to be the main bottleneck in current few-shot classification methods.

\section{Few-shot classification methods}

\begin{table*}[t]
\begin{center}
\footnotesize
\renewcommand{\arraystretch}{1.5}
\begin{tabular}{ p{3cm}|p{3cm}p{3cm}cc } 
Model & Training method & Dataset & \#params & Short name \\
\toprule
ResNet-50 \cite{he_deep_2015} & Supervised & ILSVRC2012 & 24M & \texttt{resnet50}\\
EfficientNet-B0 \cite{tan_efficientnet:_2019} & Supervised & ILSVRC2012 & 5.3M & \texttt{efficientB0}\\
EfficientNet-B7 \cite{tan_efficientnet:_2019} & Supervised & ILSVRC2012 & 66M & \texttt{efficientB7}\\
WSL (resnext101\_32x8d\_wsl) \cite{mahajan_exploring_2018} & Supervised & ILSVRC2012 + Instagram (950M images) & 88M & \texttt{wsl}\\
Robust-50~\cite{engstrom_learning_2019} & Supervised + Adv. Robustness & ILSVRC2012 & 24M & \texttt{robust50}\\
Denoising Resnet \cite{xie_feature_2018} & Supervised + Adv. Robustness & ILSVRC2012 & 70M & \texttt{denoise}\\
Local DIM \cite{bachman_learning_2019} & Unsupervised & ILSVRC2012 & 431M & \texttt{amdim}\\
\end{tabular}
\end{center}
 \caption{Pretrained deep convolutional networks used as feature extractors considered in our analysis.} 
 \label{tab:ind}
\end{table*}

The few-shot classification problem consists of a set of datasets $\mathcal{D}=\{d_0, d_1, ...\}$ where $d_n$ is an individual dataset with image and target label pairs $d_n=\{(x_0,y_0), (x_1,y_1), ...\}$. We consider each dataset $d$  split into two parts: a support set $s= \{(x_s,y_s)\}$ and a query set $q= \{(x_q,y_q)\}$. The model can access both data and labels for all examples in $s$ and is asked to predict labels for $x_q \in q$. The model is trained to minimize the cross-entropy loss:

\begin{equation} \label{eq:meta_loss}
\mathcal{L} = \mathbb{E}_{d \sim \mathcal{D}} \left[-\sum_{(x_q,y_q) \in q} y_q \log\left(f_{\theta,s}(x_q)\right)\right],
\end{equation}

where $f_{\theta,s}(x)$ stands for a parametric model which outputs the vector of inferred class probabilities. Following established nomenclature, we define $k$-shot, $N$-way classification learning as the case where the support set $s$ contains $N$ classes, and for each class we have $k$ observations. 

Following~\cite{rusu_meta-learning_2018,requeima_fast_2019}, we decompose the meta-classification model into two modules: the representation network $\phi$, and the adaptation method $a$ such that $f_{\theta,s}(x) = a_{\theta}(\phi(x_q), \{\phi(x_s), y_s\})$. In the following subsections we detail the choices made for $\phi$ and $a$ in this study.

\subsection{Pretrained models}
\label{sec:models}
\newcommand{\resnet}{resnet-50}
\newcommand{\resnetWSL}{resnet-WSL}
\newcommand{\resnetR}{resnet-R}

To extract representations from an image we use different convolutional network backbones, all based on the residual network architecture introduced in \cite{he_deep_2015}. These models consist of a stack of convolutional layers followed by a single linear layer applied to a final, flattened activation vector. For all our experiments we take this pre-logits activation vector (which we denote as $\phi(x)$) as the representation to be fed to the adaptation method.

We consider a number of recent architectures trained on the Imagenet ILSVRC2012 dataset~\cite{ILSVRC15}. For models trained on the standard supervised classification task we consider the baseline ResNet-50~\cite{he_deep_2015}, as well as the highest and lowest parameter count variants of EfficientNets~\cite{tan_efficientnet:_2019}. Additionally we consider the WSL model~\cite{mahajan_exploring_2018}, which is pretrained on an image dataset with billions of images with low quality labels before full finetuning on Imagenet. We also analyse two ResNet architectures trained with adversarial robustness~\cite{xie_feature_2018,engstrom_learning_2019} to investigate whether such representations generalize better. Finally we test the Local DIM model~\cite{bachman_learning_2019}, trained in a fully unsupervised manner on ImageNet.

\subsection{Adaptation methods}

We focus our analysis on three simple popular adaptation methods: Matching networks~\cite{vinyals_matching_2016}, Prototype networks~\cite{snell_prototypical_2017}, and Logistic regression via SGD. 

In \textit{prototype networks} the mean vector of the activations corresponding to all support examples of that class is used as a 'prototype' for that class. Class conditional probabilities are calculated as a the softmax over the similarity of a test point's activation to all 'prototypes'. Concretely, a prototype is defined as the mean vector:
$$\mu_c = \frac{1}{\#\mathcal{X}_c} \sum_{x_s \in \mathcal{X}_c} \phi(_sx)$$

where $\mathcal{X}_c$ is the set of support set inputs with label corresponding to class $c$. This is equivalent to assuming the activations for class $c$ follow a Gaussian distribution with mean $\mu_c$ and identity covariance matrix.

Assuming the labels $y_c$ of inputs belonging to class $c$ are represented by a one hot vector, the unnormalized logits $\ell$ for a query input $x_q$ are given by a similarity measure between $a_i(x_q)$ and the centroids of each class:

$$\ell\left(x_q\right) = \sum_{c = 1}^{N} \text{S}(\phi(x_q), \mu_c)\; y_c$$


where the similarity $S(\cdot,\cdot)$ can be for instance the cosine similarity, the inner product, or the negative of a distance function such as the Euclidean distance.

In \textit{matching networks}~\cite{vinyals_matching_2016} the activations for all examples in the support set are kept in memory. For a test point we query the K-nearest neighbors in activation to compute a kernel density estimate of the class conditional probabilities. The unnormalized logits are given by:
 
 $$\ell\left(x_q\right) = \sum_{(x,y)\in \mathcal{N}_K(x_q)} \text{S}\left(\phi(x_q), \phi(x)\right)\; y,$$
 
 where $\mathcal{N}_K(x_q)$ is the set of input and label pairs corresponding to the $K$ nearest neighbors of $x_q$.

One goal of our analysis is to evaluate whether the topology of the representation space impacts few-shot classification: if the representations are clustered in multiple disjoint volumes in representation space then matching networks should perform better than prototype networks, as the prototypes can only represent one single connected region of space.

We also examine the performance of simple logistic regression  on the support set's representations. For a large number of shots, we expect this method to provide an upper bound on performance as has more flexibility than similarity-based methods. For low number of shots, this method has an infinite number of solutions and therefore SGD defines an implicit form of generalization. For the experiments we use logistic regression implementation in scikit-learn~\cite{scikit-learn} with all regularization terms set to zero.

\section{Experiments}

\begin{figure*}[t]
\centering
\includegraphics[width=\textwidth]{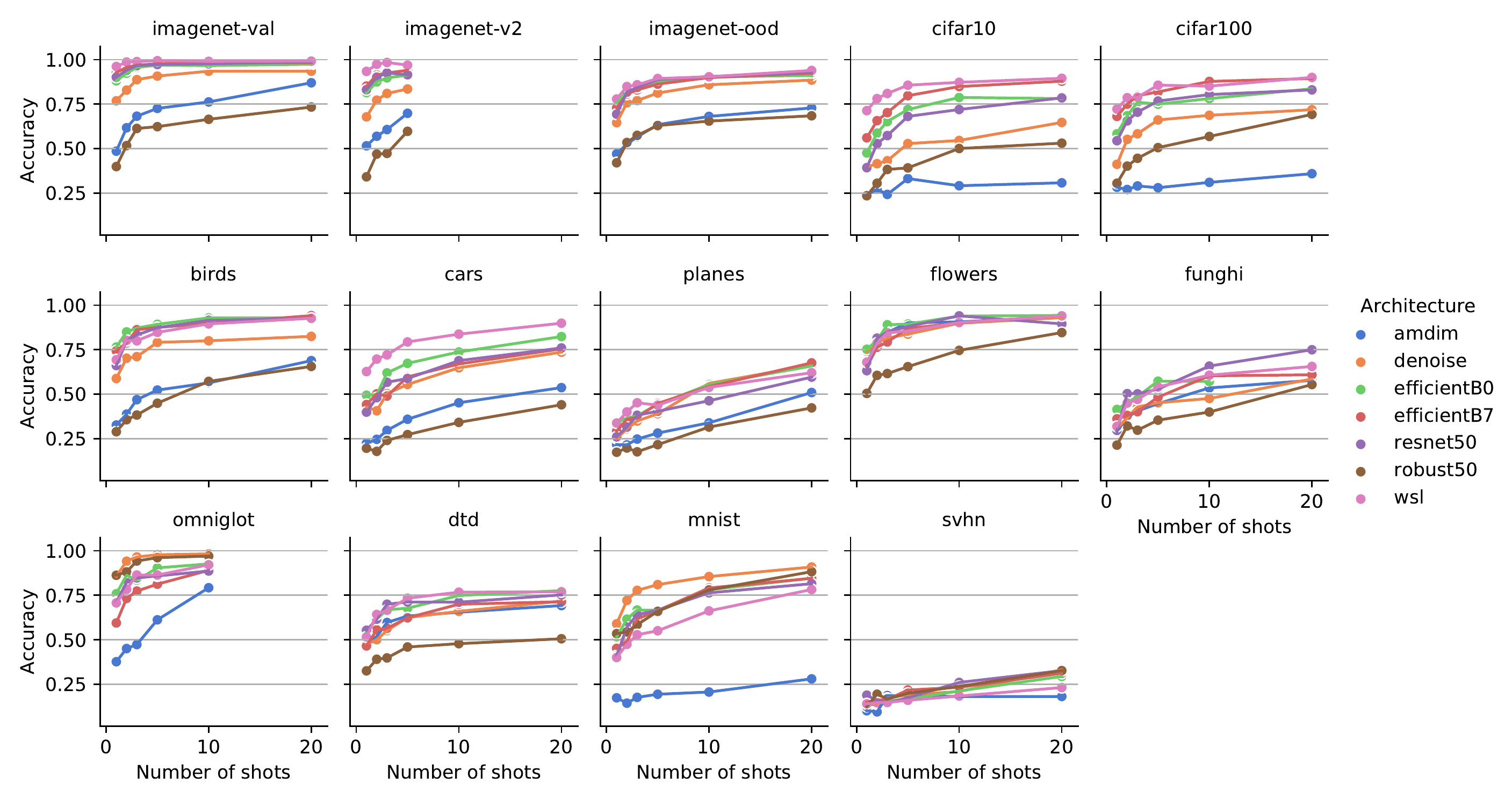}
\caption{Accuracy comparison for different datasets as a function of number of shots (elements in the support set) for 10-way classification. The accuracy value reported is the average accuracy over 25 unique episodes (where each episode is a random sample of 10 classes, from which the examples are then randomly split into a support set with $k$ examples and a query set with $\text{max}\left(z_c-k, 32\right)$ examples, with $z_c$ the total number of examples available in class $c$). Top row: we show imagenet and cifar datasets, where the best performance is obtained by networks trained on the supervised classification task. Middle row: more specialized natural image datasets where the performance decreases but supervised architectures still do better. Note that better architectures and the \texttt{wsl} network trained on more data perform better. Bottom row: out-of-distribution datasets. Here networks trained with adversarial robustness do better.}
\label{fig:fig1}
\end{figure*}

\begin{figure*}[t]
\centering
\includegraphics[width=\textwidth]{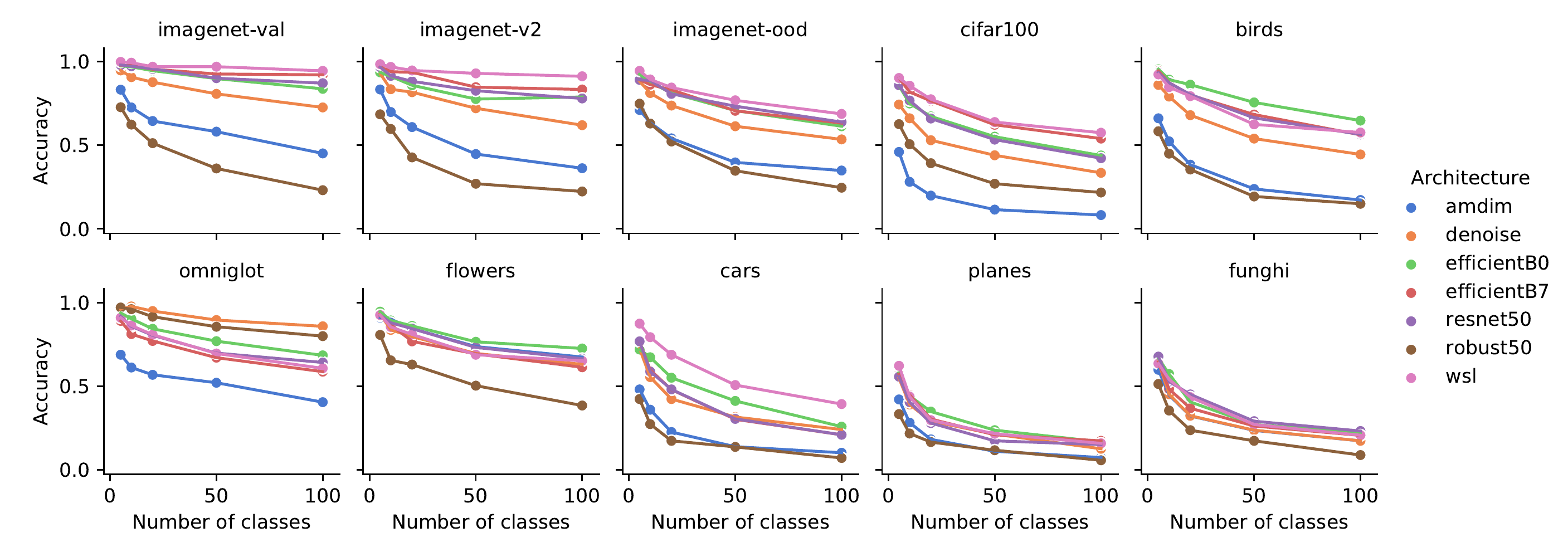}
\caption{Accuracy comparison for different datasets as a function of number of classes (elements in the support set) for 5-shot classification. The accuracy value reported is the average accuracy over 25 unique episodes (as described in the figure above). We show only results for datasets where 100 classes were available.}
\label{fig:fig4}
\end{figure*}

We evaluate the performance of all methods on 14 natural image datasets: ILSVRC2012 Imagenet validation set~\cite{ILSVRC15}; MNIST~\cite{lecun1998mnist}; Omniglot~\cite{lake_human-level_2015}; VGG flowers~\cite{Nilsback06}; FGVC-Aircraft~\cite{maji13fine-grained}; Cars~\cite{KrauseStarkDengFei-Fei_3DRR2013}; SVHN~\cite{svhn_2011}; CIFAR-10 and CIFAR-100~\cite{krizhevsky2009cifar}; DTD~\cite{cimpoi14describing}; Fungi~\cite{schroeder_fungi_2018}; Caltech Birds~\cite{WelinderEtal2010}; ImageNet-v2~\cite{recht2019imagenet}; and a subset of Imagenet classes not contained in the ILSVRC2012 1000-class set (we provide a full list of synsets in the Appendix).

For each dataset, we generate a new episode by sampling $N$ classes ($N \in \{5, 10, 20, 50, 100\}$), and create a $k$-shot support set ($k \in \{1, 2, 3, 5, 10, 20\}$) using the features generated by the network under study. Unless otherwise noted, we always use the last layer's features. The accuracy of an episode is calculated over a batch of new query datapoints (batch size 32).

\subsection{Pretrained features comparison}

\begin{figure*}[t]
\centering
\includegraphics[width=\textwidth]{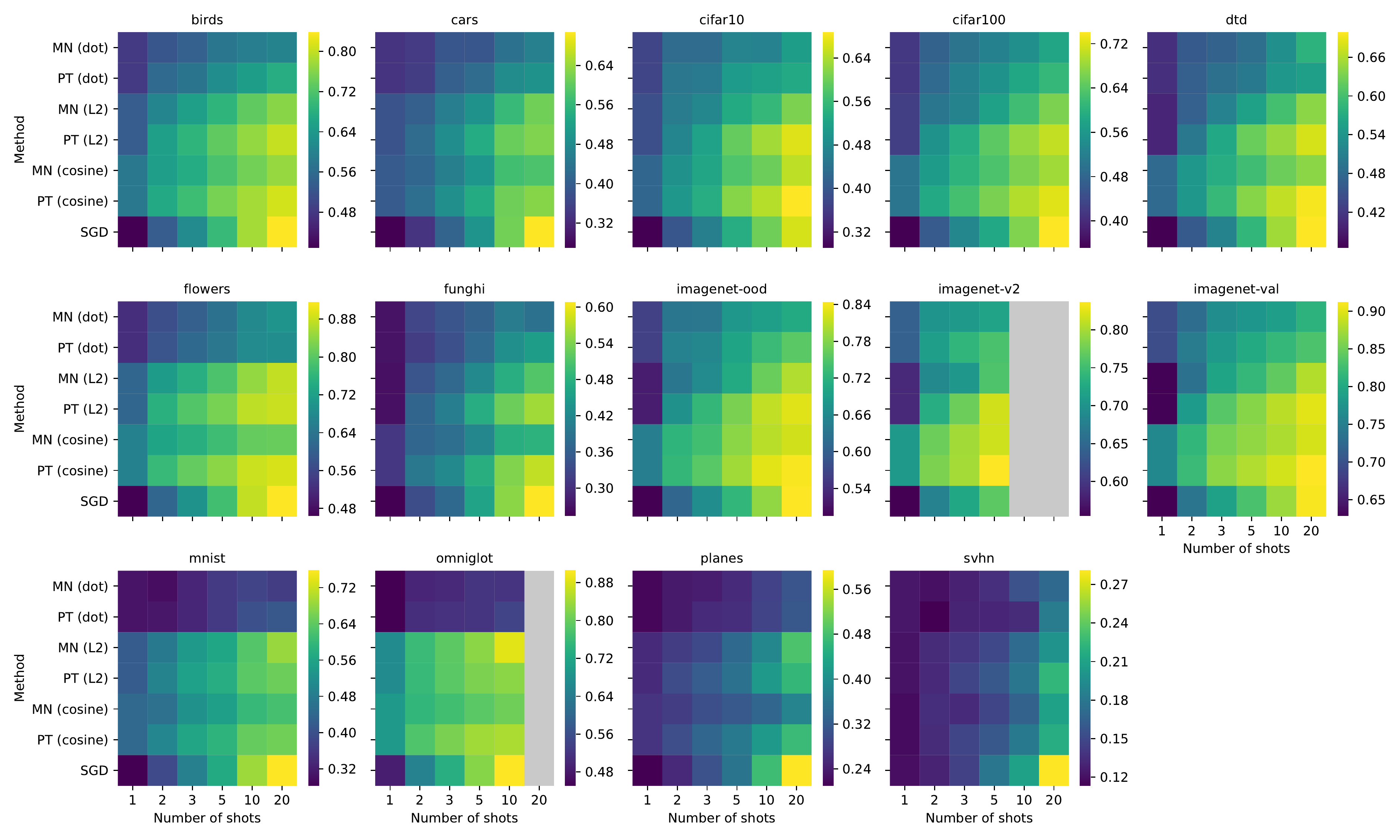}
\caption{Accuracy comparison for different datasets as a function of adaptation method and similarity function for 10-way classification. The abbreviations MN, PT and SGD are respectively used for Matching Networks, Prototype Networks, and Logistic Regression with Stochastic Gradient Descent. Light grey values represent missing data (when the dataset does not contain enough examples per class for that number of shots). The accuracy value reported is the average accuracy over 25 unique episodes (as described in Fig.\ref{fig:fig1}).}
\label{fig:fig2}
\end{figure*}

Firstly we seek to quantify how the features calculated by the various pretrained backbones under consideration affect the final performance of the few-shot classification task across a range of number of shots and number of classes under consideration. Are certain backbones universally better than others? Do unsupervised models and models trained with adversarial robustness generalize better?

In Figure \ref{fig:fig1} we compare the 10-way classification performance across all architectures under consideration for different datasets as a function of number of shots (elements in the support set). To focus on the feature quality, we report the maximum accuracy achieved over all adaptation methods for that particular architecture and number of shots. In the next section we will break the results down by adaptation method.

Firstly, we observe that supervised classification architectures all do very well on the three datasets derived from imagenet, as would be expected since there is little distributional shift. As expected from the results in~\cite{recht2019imagenet}, there is a small drop in performance for ImageNet-v2, and a larger performance drop for the ImageNet dataset with unseen classes. Performance on the cifar datasets is also high, in spite of a very different image resolution (we upscale all images where the resolution does not match the original model's input resolution).

The same architectures also perform well in various natural image based datasets such as birds, cars, flowers. For these datasets we empirically observe that architectures with better final performance on the classification task perform better. In particular the \texttt{wsl} network, pre-trained on much more data, stands out.

Out-of-distribution datasets such as omniglot, dtd, mnist and svhn present a different picture. Not only does the performance degrade, networks trained with adversarial robustness do better than their non-robust counterparts. The architecture trained in a fully unsupervised manner does not generalize better, in spite of its features not having been tuned in a class-specific manner by a supervised loss.

In Figure \ref{fig:fig4} we plot the same quantity as above, but now as a function of the number of classes for a fixed number of shots (5 shot per class). We can confirm with this plot that the relative ranking between models stays roughly constant as they degrade in performance with increasing number of classes. In fact, performance for a low number of classes seems to be predictive of final performance for high number of classes across architectures and datasets, with higher drops for lower initial accuracy numbers.

\subsection{Adaptation method comparison}

Next we quantify the effect of the adaptation method on the final performance of few-shot classification when used with pre-trained models. Since pretrained models were not trained with few-shot classification in mind, conclusions reached by previous studies~\cite{chen_closer_2019,triantafillou_meta-dataset:_2019,snell_prototypical_2017} may not hold true.

Specifically we are interested in the following questions: if a model is trained in a supervised manner with a final softmax layer, the representations have no reason to cluster together in the sense of the euclidean distance, as what matters to produce discriminative logits is the dot product of the representation vector with a given entry in the weights matrix of the last linear layer. Therefore, the averaging done by prototype networks could be detrimental, and the use of $L_2$ distance might not be optimal in terms of determining proximity in this space.

In Figure \ref{fig:fig2} we compare the accuracy different datasets as a function of adaptation method and similarity function for 10-way classification. The accuracy is reported as an average over all architectures under consideration. From these heatmaps we can observe that contrary to our expectation Prototype Networks seems to consistently beat Matching Networks. On the other hand we do find that cosine similarity is consistently superior to $L_2$ distance as we expected. This is different to the results originally reported in Prototype Networks, where $L_2$ worked better for a convolutional backbone pretrained from scratch.

Furthermore we observe that the unnormalized dot product tend to perform worse than the other similarity measures. We hypothesise this is due to the fact that there are no learnable parameters in these adaptation methods and therefore an unnormalized method suffers if there are scale differences between the average features of different classes. We leave this investigation to futre work.

Finally we observe that Logistic regression with SGD is the best performer for 10 shots and above, while being significantly worse than other adaptation methods for a small number of shots. This agrees with the intuition of the early papers introducing few-shot classification~\cite{vinyals_matching_2016}, which propose their methods only in the 5-shot and below regime. 

This suggests that at least in the case of pretrained models there is a clear hierarchy of which methods are best as a function of number of shots: for 1-5 shots, methods such as Prototype and Matching networks will work best; for over 10 shots, retraining of the last layer will do better; and finally for high number of examples per class fine-tuning becomes a viable option~\cite{kornblith_better_2018}.

\begin{figure*}
\centering
\includegraphics[width=0.8\textwidth]{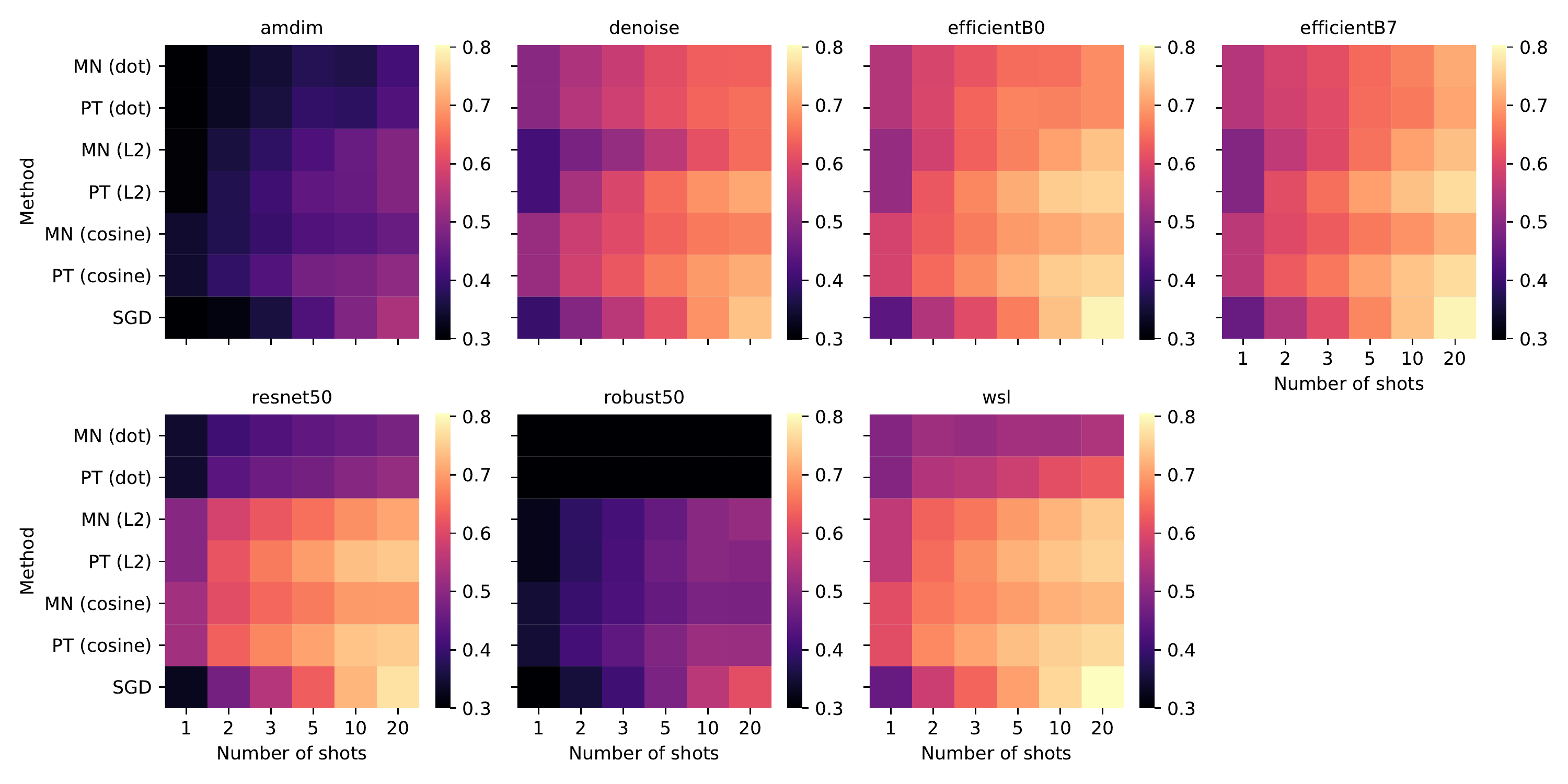}
\caption{Accuracy comparison for different pretrained backbones as a function of adaptation method and similarity function for 10-way classification. The abbreviations MN, PT and SGD are respectively used for Matching Networks, Prototype Networks, and Logistic Regression with Stochastic Gradient Descent. The accuracy value reported is the average accuracy over all datasets.}
\label{fig:fig3}
\end{figure*}

In Figure \ref{fig:fig3} we break down the accuracy per pre-trained architecture. We find that there are no significant differences to the analysis described above, with Prototype Networks with cosine similarity generally performing best among similarity-based methods. Unnormalized dot product still exhibits the worst performance, although with EfficientNets its performance drop is not so significant. Overall the most robust architecture across all datasets appears to be \texttt{efficientB7}, while the individual highest performer for the best hyperparameter choices is \texttt{wsl}.

\section{Conclusions}

In this paper we systematically investigate the performance of pretrained models as backbones to calculate feature representations for few-shot image classification tasks. We tested their accuracy when used as starting point for popular similarity-based few-shot classification methods across a range of datasets.

Encouragingly we find that these models can provide very high accuracy for a number of few-shot classification tasks. We found that models pre-trained on a supervised classification task on the ImageNet dataset transfer well to other natural image datasets, but poorly to other types of image (e.g. MNIST, SVHN). The more data models are trained on, the better their transfer performance.

Contrary to our expectation, models trained with adversarial robustness or trained with an unsupervised loss do not seem to outperform models trained on the popular 1000-class ILSVRC2012 ImageNet classification task. However, the models tested here do not achieve the same performance on the original ILSVRC2012 test set as the purely supervised models, so their performance should be revisited when the field matures and performance improves.

We also found that some best practices on few-shot classification do not transfer to the use of pre-trained models (e.g. we find that the cosine similarity provides better performance than $L_2$ distance, contrary to what is the case with end-to-end trained few-shot classification models).

We hope our empirical investigation will spur the use of pretrained models in applied few-shot classification and online learning tasks, as they can provide excellent performance with very little training costs. All convolutional backbones we experimented with have publicly available weights and code implementations and we thank all the respective authors for releasing their code for experimentation.

\bibliography{example_paper}
\bibliographystyle{plain}

\end{document}